\documentclass[11pt]{article}

\usepackage{acl}

\usepackage{times}
\usepackage{latexsym}

\usepackage[T1]{fontenc}

\usepackage[utf8]{inputenc}

\usepackage{microtype}

\usepackage{inconsolata}

\usepackage{graphicx}

\usepackage{booktabs}
\usepackage[most]{tcolorbox}
\usepackage{subfigure}
\usepackage{amssymb}
\usepackage[capitalize,noabbrev]{cleveref}
\usepackage{makecell}
\usepackage{float}

\newcommand\ours{\textsc{LongReasonArena}}
\newcommand\myline{ }

\title{\ours{}: A Long Reasoning Benchmark \\ for Large Language Models}

\author{Jiayu Ding$^1$\footnotemark[1] \quad Shuming Ma$^2$ \quad Lei Cui$^2$ \quad
Nanning Zheng$^1$\footnotemark[2] \quad Furu Wei$^2$\footnotemark[2] \\
$^1$ State Key Laboratory of Human-Machine Hybrid Augmented Intelligence, \\
Institute of Artificial Intelligence and Robotics, Xi'an Jiaotong University \\
$^2$ Microsoft Research Asia \\
\texttt{\fontsize{8}{9}\selectfont nachewigkeit@stu.xjtu.edu.cn, nnzheng@mail.xjtu.edu.cn, \{lecu,fuwei\}@microsoft.com}}

\begin{document}
\maketitle
\footnotetext[1]{Work done during the internship at Microsoft Research Asia.}
\footnotetext[2]{Corresponding authors.}
\begin{abstract}

Existing long-context benchmarks for Large Language Models (LLMs) focus on evaluating comprehension of long inputs, while overlooking the evaluation of long reasoning abilities. To address this gap, we introduce \ours{}, a benchmark specifically designed to assess the long reasoning capabilities of LLMs. Our tasks require models to solve problems by executing multi-step algorithms that reflect key aspects of long reasoning, such as retrieval and backtracking. By controlling the inputs, the required reasoning length can be arbitrarily scaled, reaching up to 1 million tokens of reasoning for the most challenging tasks. 
Extensive evaluation results demonstrate that \ours{} presents a significant challenge for both open-source and proprietary LLMs. For instance, even GPT-5 achieves only 27.8\% accuracy on Level 3. Further analysis also reveals that the accuracy exhibits a linear decline with respect to the logarithm of the expected number of reasoning steps. Our code and data are available at \url{https://github.com/LongReasonArena/LongReasonArena}.

\end{abstract}

\section{Introduction}

\begin{figure*}[t]
    \centering
    \includegraphics[width=\textwidth]{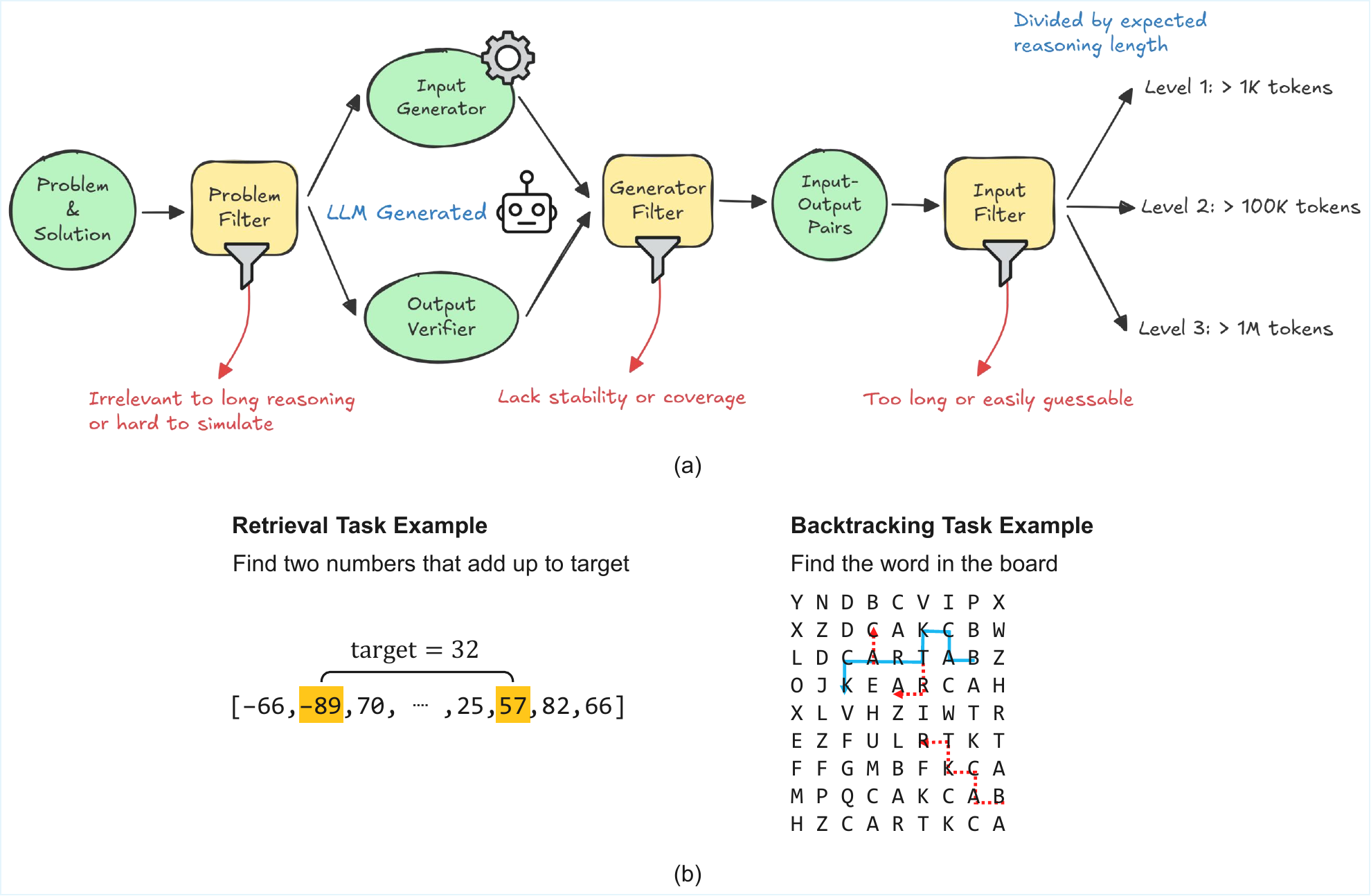}
    \caption{(a) The construction process of \ours{} (b) Algorithmic problems abstract key elements of long reasoning, including retrieval, backtracking, and other essential cognitive operations.}
\end{figure*}

Recently, Large Language Models (LLMs) have extended the concept of scaling laws from scaling up train-time compute to scaling up test-time compute, achieving remarkable improvements on tasks such as mathematics, programming, and question answering \cite{o1, r1, qwq}. To enable test-time scaling, Large Language Models are trained via reinforcement learning to generate extended chain-of-thought that spans thousands of tokens. As a result, their reasoning lengths significantly exceed those of previous approaches \cite{cot, tot}, introducing new challenges and considerations for evaluation.

Existing long-context benchmarks focus mainly on long input, such as LongBench \cite{longbench} and Needle-in-a-Haystack  \cite{needle}. However, long reasoning differs fundamentally from long input. In the case of long input, the model only needs to passively receive and comprehend information. In contrast, long reasoning requires the model to actively generate, structure, and self-correct its output. Some recent benchmarks have begun to evaluate long-form generation, such as LongGenBench \cite{longgenbench_liu} and GSM-Infinite \cite{gsminf}, but these tasks are relatively simple and remain far from reflecting the complexity of long reasoning processes.

To better evaluate the long reasoning capabilities of the models, we propose \ours{}, which assesses performance by requiring the models to execute algorithms to solve problems. Executing algorithms inherently requires the model to perform long reasoning, involving a large number of algorithmic steps. The number of required steps and the corresponding reasoning length can be controlled by varying the input, which allows us to arbitrarily scale the required reasoning length. During algorithm execution, some essential capabilities required for long reasoning are put to the test, such as retrieval and backtracking. Research in cognitive science posits that memory and backtracking play important roles in human reasoning \cite{human_problem_solving, soar, working_memory, cognitive_offloading}. Empirical observations of LLMs' long reasoning confirm a similar dependency: models must retrieve prior intermediate steps to maintain coherence and employ backtracking to correct errors or switch strategies. These reasoning dynamics are mirrored in the domain of algorithm execution. For instance, algorithmic procedures like depth-first search inherently test a model's ability to perform extensive backtracking. Likewise, the manipulation of data structures—requiring dynamic updates such as insertion and deletion—poses a comparable challenge to the model's ability to manage and retrieve information effectively.

To construct the benchmark, we collect algorithmic problems and their corresponding solution code from LeetCode. For each problem, we use GPT-5 \cite{gpt5_release} to generate an input generator function and an output verifier. The input generator is required to stably produce valid inputs and ensure sufficient coverage over the solution code, which enables the generation of an unlimited number of diverse inputs. Problems lacking a qualified generator are excluded from the benchmark. We further apply model-based filtering with GPT-5 and Claude Opus 4.5 \cite{opus45_release} to exclude problems that models can solve through shortcuts or guessing rather than long reasoning.

We categorize tasks into different difficulty levels based on the expected number of reasoning steps. To estimate the reasoning steps required to solve each problem, we measure the number of execution lines when running the solution code on the given input. \ours{} is divided into three levels: Level 1, 2, and 3 correspond to approximately 1K, 100K, and 1M tokens of reasoning, respectively. \ours{} poses a significant challenge for both open-source and proprietary LLMs. For example, even GPT-5 achieves only 27.8\% accuracy on Level 3.

We evaluate fourteen models on \ours{}, including reasoning and non-reasoning models of various sizes. Based on the analysis of the evaluation results, we reveal several limitations of current reasoning models. First, accuracy decreases as the expected number of reasoning steps increases, following a linear relationship with the logarithm of the step count. Second, models struggle with basic reasoning operations such as retrieval and backtracking: for instance, although models perform well on long-input retrieval tasks, they still often fail at retrieval operations within long reasoning chains, such as checking whether a number appears. 

The contributions of this paper are summarized as follows:

\begin{itemize}
    \item We propose \ours{}, a benchmark for evaluating long reasoning by algorithmic execution, where the required reasoning length can be arbitrarily scaled (up to 1M tokens) through input control. Our tasks reflect key aspects of long reasoning such as retrieval and backtracking.
    \item We evaluate fourteen models on \ours{}. Our analysis reveals that model accuracy declines linearly with the logarithm of the expected reasoning steps, and that current models still struggle with basic retrieval and backtracking operations.
    \item We publicly release the code and data for \ours{} to support reproducible evaluation and facilitate future research on long reasoning.
\end{itemize}

\section{Related Work}
\paragraph{Long Input Benchmarks} Existing long-context benchmarks primarily focus on evaluating LLMs' ability to comprehend long inputs, such as the ability to retrieve and summarize \cite{longbench,leval,needle,ruler,infty}. Some papers evaluate models’ reasoning ability using a retrieve-then-reason approach \cite{alr,longreason,babilong,docpuzzle}, but the challenge still primarily lies in extracting information from long inputs rather than performing long reasoning. In contrast, \ours{} explicitly requires models to engage in a long reasoning process to solve the task.

\paragraph{Long Generation Benchmarks} Recently, several studies have begun to explore the evaluation of models' long generation capabilities. Some long-generation benchmarks \cite{longgenbench_liu,longgenbench_wu} construct tasks based on relatively simple and repetitive patterns, with subtasks that are independent of one another. GSM-Infinite \cite{gsminf} abstracts GSM-8K problems into computational graphs with controllable reasoning complexity and noise level, but its problem space is limited to composite arithmetic operations. LR$^2$Bench \cite{lr2bench} evaluates long-chain reflective reasoning, but contains only six problems, all restricted to constraint satisfaction. L0-Bench \cite{l0bench} evaluates procedural correctness on generated simple programs, but it imposes strict constraints, prohibiting operations like $*$, $\%$, and $<$. This severely restricts its ability to simulate complex reasoning structures, such as recursion or backtracking. Overall, these benchmarks fall short in capturing the complexity of long reasoning processes. In contrast, \ours{} encompasses a more diverse set of tasks, enabling a broader evaluation across a wider range of reasoning challenges. Other benchmarks, such as TurnBench-MS \cite{turnbenchms}, focus on multi-turn interactive reasoning, while \ours{} evaluates whether models can complete continuous long reasoning within a single turn.

\paragraph{Code Execution} Some recent works leverage LLMs as code executors, tasking models with simulating program execution or predicting program behavior. CodeI/O \cite{codeio} prompts DeepSeek-V2.5 \cite{deepseekv2} to perform input/output prediction, and uses the generated reasoning traces as training data for other models. Lyu et al.\cite{executor} curated 200 code snippets with input-output examples to evaluate line-by-line execution, while REval \cite{reval} evaluates runtime behavior prediction. These benchmarks are valuable for studying program simulation, but their tasks are relatively simple and insufficiently challenging for evaluating strong reasoning models on long reasoning. Non-reasoning models already perform well on these benchmarks. For example, GPT-4-Turbo achieves 82.6\% output-prediction accuracy on REval. In contrast, \ours{} is explicitly designed to evaluate long reasoning capabilities, presenting tasks that are challenging even for state-of-the-art reasoning models.

\paragraph{Program Synthesis} The program synthesis capability of LLMs has attracted widespread attention. Works such as HumanEval\cite{humaneval}, MBPP\cite{mbpp}, and APPS\cite{apps} have evaluated the program generation ability of LLMs. LiveCodeBench\cite{livecodebench} further addresses the issue of data contamination. However, these benchmarks are not designed to directly evaluate a model’s long reasoning ability, making it difficult to estimate the number of reasoning steps required for solving a problem. For instance, if a model has encountered similar code snippets during training, the reasoning difficulty can be significantly reduced. In contrast, our benchmark requires the model to execute every step correctly in order to arrive at the correct answer, with both controllable and scalable task complexity.

\begin{table*}[t]
\centering
\begin{tabular}{@{}lccc@{}}
\toprule
                                & Level 1 & Level 2 & Level 3 \\ \midrule
\# Problems                     & 322    & 390     & 368    \\
\# Samples                      & 696    & 908     & 769   \\
Median execution line count     & 1,197     & 30,683   & 183,628 \\
90th percentile of execution line count & 7,045 & 79,184 & 544,348 \\
Median input length             & 66     & 680    & 2,702    \\
90th percentile of input length & 766    & 8,110   & 23,494  \\
\bottomrule
\end{tabular}
\caption{The statistics of datasets at different levels.}
\label{tab:stat}
\end{table*}

\section{Benchmark Construction}
\label{sec:construction}

\subsection{Problem Selection}

Our goal is to select diverse problems that reflect key characteristics of long reasoning processes. Our problems are selected from LeetCode
, which provides a broad coverage of algorithmic topics and offers detailed information for each problem, including tags, solutions, and input constraints. We first filter the dataset by tags, excluding problems such as “Database” and “Randomized” that primarily test implementation skills or are hard to simulate with LLMs. To facilitate input generation and output verification, we further exclude problems that have no return value, involve custom classes in type annotations (e.g., "TreeNode"), or involve floating-point numbers in type annotations, which are prone to precision issues.

Backtracking and memory are two fundamental components of long reasoning. Backtracking is a phenomenon unique to long reasoning and does not arise in tasks that only involve processing long inputs. It reflects the model's ability to explore multiple reasoning paths and reconsider its approach by abandoning the current trajectory and trying alternatives. Memory, on the other hand, is essential for both long reasoning and long input processing. However, in the context of reasoning, memory must be dynamically updated to incorporate new intermediate conclusions and revise previous errors as the thought process unfolds. 

To better evaluate the model’s backtracking and memory capabilities, we select a subset of problems as core problems and sample more inputs for each. These core problems are oversampled for targeted analysis, while the overall dataset remains diverse and covers a wide range of algorithmic tags. For backtracking, we choose core problems tagged with “Depth-First Search” or “Backtracking.” For memory, we choose core problems tagged with “Dynamic Programming” or “Breadth-First Search,” as dynamic programming typically involves maintaining and updating a value table, while breadth-first search requires managing a dynamically evolving queue. Analysis of LLM reasoning traces across all problems confirms that these selected core problems indeed elicit more frequent backtracking and memory manipulation behaviors (details provided in \cref{sec:core_problems}).

\subsection{Sample Generation}

\paragraph{Input Generator} To facilitate the generation of inputs with unbounded quantity and complexity, we leverage GPT-5 \cite{gpt5_release} to generate an input generator function for each problem. We define two criteria to evaluate the effectiveness of the input generator: stability and coverage. Stability requires that the generator consistently produces valid inputs without triggering errors in the solution function, and code coverage is a standard metric in software engineering, defined as the percentage of source code executed by a test suite. In our validation pipeline, the source code is the ground-truth solution code for a problem, and the test suite consists of multiple distinct inputs produced by the generator. Therefore, higher coverage indicates that the generated inputs explore more edge cases and logical branches of the solution code. In practice, we consider a generator to be qualified if it can consecutively generate ten valid inputs and achieve over 90\% code coverage. Problems without a qualified generator are excluded from the benchmark. Input generator functions generated by GPT-5 meet the qualification criteria for 94\% of collected LeetCode problems, which is sufficient for our purposes.

\paragraph{Difficulty Level} We categorize data samples (problem-input pairs) into different difficulty levels based on the number of execution lines measured when running the solution code on the given input. This metric serves as a proxy for the number of steps required to solve the problem. Specifically, samples with between $10^2$ and $10^4$ execution lines are labeled as Level 1, those with between $10^4$ and $10^5$ lines as Level 2, and those with between $10^5$ and $10^6$ lines as Level 3. Although models may occasionally solve certain samples using fewer steps by guessing or heuristic methods, the average number of reasoning steps at each level remains proportional to the average number of execution lines.

We estimate that each execution line typically requires at least 10 tokens to complete, such as performing a comparison or computation command. Accordingly, solving all Level 2 tasks without guessing would require a model to process at least 100K tokens of reasoning per sample; for Level 3, this requirement increases to at least 1M tokens per sample. These demands far exceed the current capabilities of existing reasoning models. \ours{} is designed to challenge and guide the development of reasoning models, and to serve as a benchmark for future models capable of handling longer reasoning processes.

\paragraph{Input Length Constraint} To ensure that the primary challenge of the task lies in long reasoning rather than long input, we retain only those samples whose input length is within 32K tokens. Dataset statistics are shown in \cref{tab:stat}. Even for the Level 3 subset, the median input length is only 2,702 tokens, and the 90th percentile remains below 24K tokens. Compared to the typical 128K context length supported by current models, this is relatively small, ensuring that the core difficulty of our task does not arise from handling long inputs.

\paragraph{Model-based Filtering}
Beyond controlling difficulty by execution length, we further apply model-based filtering to remove problems that are already reliably solvable by frontier reasoning models. Execution lines provide only an approximate estimate of reasoning difficulty, since some samples can be solved by shortcuts or simple guesses. For example, in a task that asks for all triplets in an array that sum to zero, a model must search for the triplets when they exist; but when no triplet exists, it may terminate early and guess ``no solution'' after only a few attempts. Such samples do not effectively test long reasoning, even if the reference solution executes many lines.

We therefore apply model-based filtering. For each candidate problem, we generate five Level 3 samples and evaluate them with GPT-5 and Claude Opus 4.5 \cite{opus45_release}. If either model correctly solves all five samples, we discard the problem. Using two model families reduces model-specific bias and avoids constructing a benchmark around the weaknesses of a single system.

\paragraph{Output Verification}

During evaluation, models are prompted to enclose their final answer within \verb|\boxed{}|. Any formatting errors or incorrect answers are treated as incorrect. The reference answer for each sample is obtained by executing the solution code. However, some problems permit multiple correct outputs (e.g., when all combinations or permutations are to be returned in any order). To account for such cases, we use GPT-5 to determine whether the problem is input-dependent in this regard, and to generate code that verifies whether the model's generated answer is equivalent to the reference answer.

\section{Evaluation}
\label{sec:evaluation}

\begin{table*}[t]
\centering
\begin{tabular}{@{}lcccc@{}}
\toprule
                             & Reasoning Model & Level 1   & Level 2   & Level 3 \\
\midrule
GPT-5               & $\checkmark$ & 84.2        & 48.3    & 27.8 \\
Claude Opus 4.5             & $\checkmark$ & 66.1      & 29.1   & 17.6      \\
o3                           & $\checkmark$ & 65.4        & 36.6    & 22.4 \\
o1                           & $\checkmark$ & 62.8        & 32.0    & 20.0 \\
Qwen3-32B & $\checkmark$ & 56.3      & 20.3    & 13.5    \\
QwQ                          & $\checkmark$ &  56.2     &  20.6   &  13.4  \\
Qwen3-14B & $\checkmark$ & 55.6      & 20.0    & 11.4    \\
Claude Sonnet 4.5            & $\checkmark$ & 54.3      & 18.9   & 10.9      \\
Qwen3-8B & $\checkmark$ & 50.4      & 14.9    & 8.5    \\
Qwen3-4B & $\checkmark$ & 47.3      & 13.0    & 8.3    \\
DeepSeek-R1                  & $\checkmark$ & 42.2      & 16.4    & 9.9     \\
DeepSeek-R1-Distill-Qwen-32B & $\checkmark$ &  42.1     &  11.2   &  7.7   \\
GPT-4o                       & $\times$ & 25.0        & 7.3     & 4.3       \\
Llama 3.3 70B Instruct       & $\times$ & 22.7        & 8.7    &  5.9      \\
\bottomrule
\end{tabular}
\caption{Evaluation results of 14 models across difficulty levels.}
\label{tab:leaderboard}
\end{table*}

\subsection{Setup \& Results}
\label{subsec:setup}
We evaluate 14 models \cite{gpt5_release,o3,o1,gpt4o,r1,qwen3,qwq,opus45_release,sonnet45,llama}, including 8 open-source models and 6 proprietary models. For all open-source models except DeepSeek-R1 (due to the difficulty of deploying DeepSeek-R1), we perform inference using vLLM \cite{vllm} on 8 NVIDIA A100 GPUs. For the proprietary models and DeepSeek-R1, we use API calls. Detailed inference settings are provided in \cref{sec:param}.

The evaluation results are shown in \cref{tab:leaderboard}. Overall, large reasoning models significantly outperform non-reasoning models. However, for Level 2 and Level 3 tasks, state-of-the-art reasoning models still struggle. For instance, even GPT-5 reaches only 27.8\% accuracy at Level 3. We also evaluated human performance on \ours{}, finding that it remains challenging for humans as well (see \cref{sec:human} for details).

\subsection{Impact of Reasoning and Input Complexity on Accuracy}
\begin{figure*}[t]
    \centering
    \subfigure[]{\label{subfig:steps}
    \includegraphics[width=0.45\textwidth]{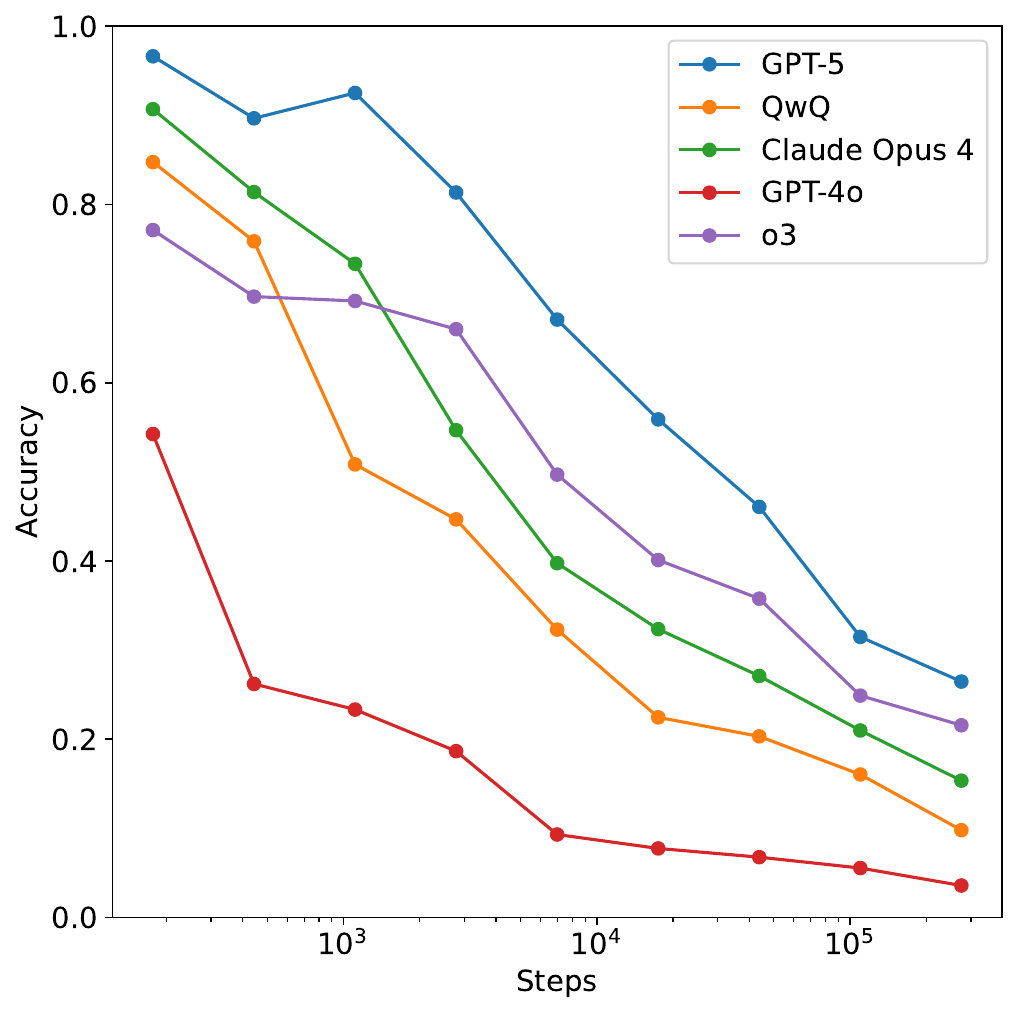}
    }
    \subfigure[]{
    \includegraphics[width=0.45\textwidth]{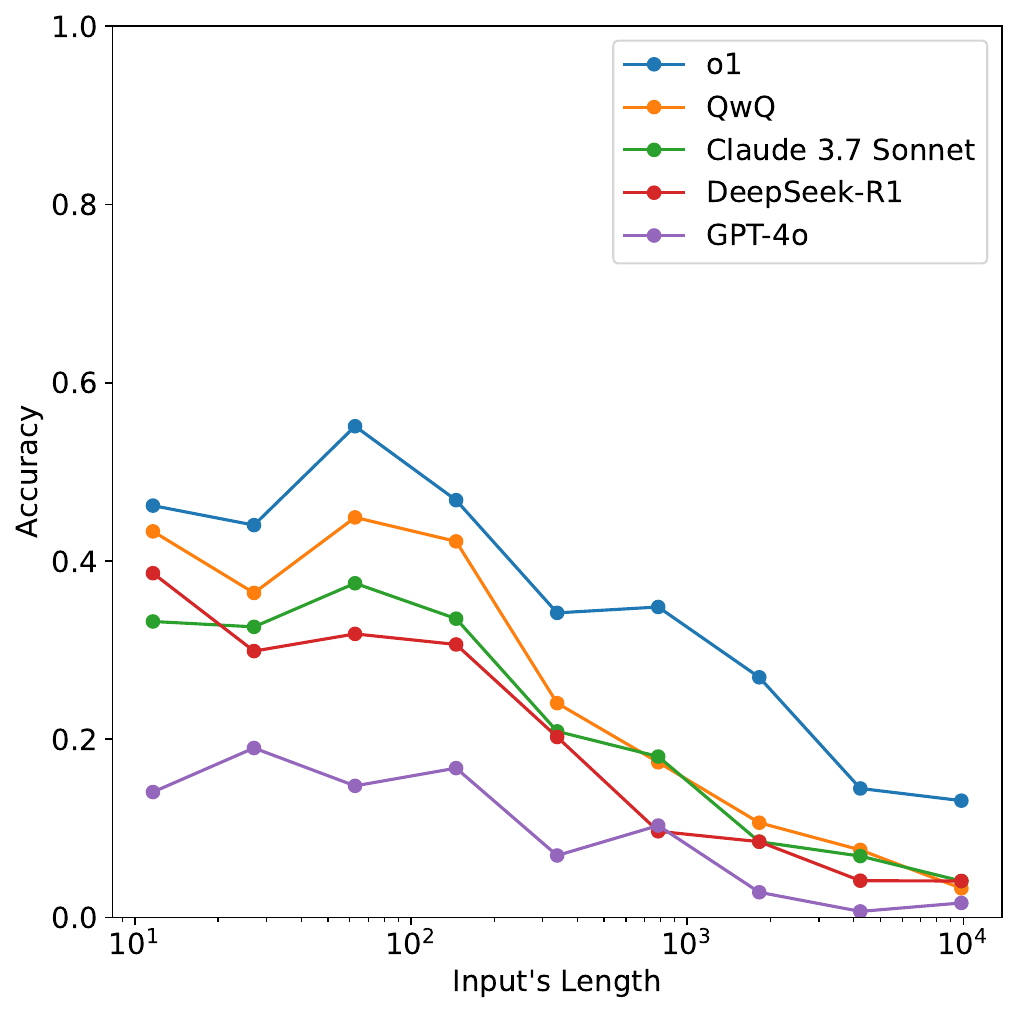}
    }
    \caption{Model accuracy with respect to (a) expected number of reasoning steps and (b) input length. Accuracy consistently decreases with both increasing reasoning and input difficulty. A strong log-linear relationship is observed with respect to the expected number of reasoning steps across all reasoning models.}
    \label{fig:trend}
\end{figure*}

We analyze the impact of expected number of reasoning steps and input length on model accuracy, as shown in \cref{fig:trend}. Accuracy decreases as the expected number of reasoning steps increases, following a strong linear trend with respect to the logarithm of steps. For reasoning models like GPT-5 and Claude Opus 4.5, $R^2>0.96$ (detailed results in \cref{sec:linear}). This indicates that our estimation of the required reasoning steps serves as an effective proxy for task difficulty.

With respect to input length, model accuracy varies only slightly when inputs are short, suggesting limited sensitivity to small-scale inputs. However, as the input length increases beyond approximately $10^2$ tokens, accuracy drops sharply.

Across both trends, reasoning models exhibit similar patterns of decline, while the non-reasoning model GPT-4o performs consistently worse. This consistency across diverse reasoning models highlights that the observed trends are not model-specific but rather reflect a general property of reasoning models.

\subsection{Impact of Inference-Time Budgets on Accuracy}

We further investigate how inference-time budgets affect model performance by varying reasoning effort and context length. Specifically, we evaluate GPT-5 with the high reasoning effort setting, and extend the context window of Qwen3-8B from 32K to 128K using YaRN \cite{yarn} following the official guidelines.

As shown in \cref{tab:context_effort}, increasing the reasoning effort of GPT-5 improves accuracy across all three levels, with gains of 2.4, 1.5, and 5.8 percentage points on Levels 1, 2, and 3, respectively. Despite these gains, GPT-5 still struggles on harder tasks. In contrast, extending the context length of Qwen3-8B from 32K to 128K changes accuracy by only $-0.3$, $-0.2$, and $+0.3$ percentage points across the three levels. This contrast suggests that performance is primarily constrained by the model's long reasoning ability rather than the context window itself. When a model lacks sufficient capability to sustain and organize long reasoning chains, simply increasing the available context length does not enable it to solve substantially harder instances.

\begin{table}[H]
\centering
\begin{tabular}{@{}lccc@{}}
\toprule
Model & Level 1 & Level 2 & Level 3 \\
\midrule
GPT-5 (medium) & 84.2 & 48.3 & 27.8 \\
GPT-5 (high) & 86.6 & 49.8 & 33.6 \\
Qwen3-8B (32K) & 50.4 & 14.9 & 8.5 \\
Qwen3-8B (128K) & 50.1 & 14.7 & 8.8 \\
\bottomrule
\end{tabular}
\caption{Impact of reasoning effort and context length on model accuracy.}
\label{tab:context_effort}
\end{table}

\subsection{Connections to Real-world Tasks}
\label{subsec:realworld}
Although \ours{} evaluates models through controlled algorithmic execution, the resulting model ranking is broadly consistent with representative real-world reasoning tasks. As shown in \cref{tab:realworld}, models that perform better on AIME 2025 \cite{aime2025}, GPQA Diamond \cite{gpqa}, and LiveCodeBench \cite{livecodebench} also tend to rank higher on \ours{}. This alignment suggests that the capabilities tested by \ours{}, including sustained multi-step reasoning, state tracking, retrieval, and backtracking, are relevant to practical reasoning-intensive applications.

\begin{table*}[t]
\centering
\begin{tabular}{@{}lcccc@{}}
\toprule
Model & AIME 2025 & GPQA Diamond & LiveCodeBench & \ours{} \\
\midrule
GPT-5 & 94.6 & 85.7 & 84.6 & 84.2 \\
Claude Opus 4.5 & 91.3 & 87.0 & 87.1 & 66.1 \\
o3 & 88.9 & 83.3 & 80.8 & 65.4 \\
Qwen3-32B & 72.9 & 68.4 & 65.7 & 56.3 \\
DeepSeek-R1 & 70.0 & 71.5 & 63.5 & 42.2 \\
\bottomrule
\end{tabular}
\caption{Comparison between model performance on real-world reasoning tasks and \ours{}. Scores on AIME 2025, GPQA Diamond, and LiveCodeBench are collected from public sources \cite{gpt5_release,opus45_release,qwen3,r1,artificial_analysis}. The \ours{} column reports Level 1 accuracy.}
\label{tab:realworld}
\end{table*}

\subsection{Analysis of Retrieval and Backtracking Failures}

\begin{figure*}[t]
    \centering
    \subfigure[Two Sum]{
        \label{fig:twosum}
        \includegraphics[width=0.4\textwidth]{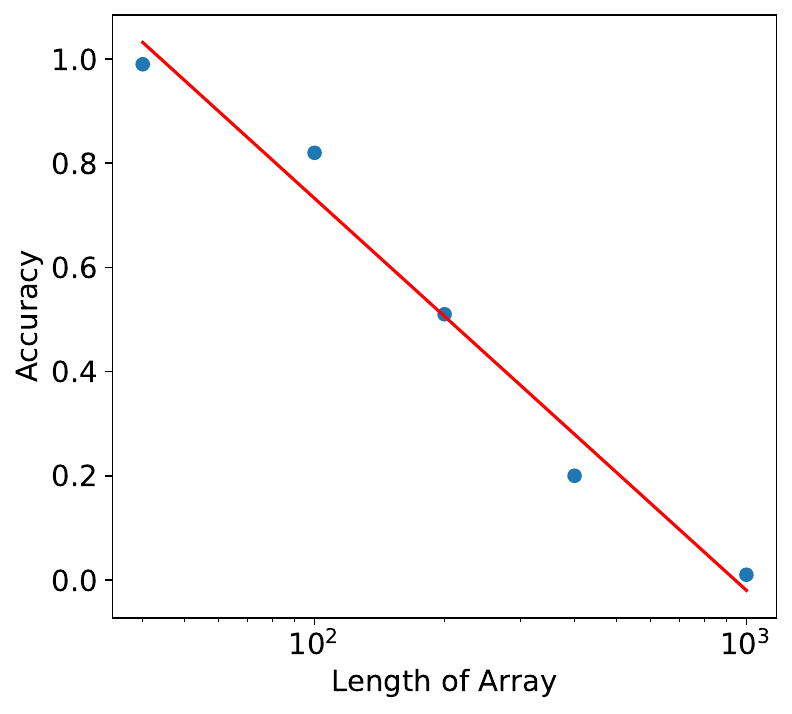}
    }
    \subfigure[Word Search]{
        \label{fig:word}
        \includegraphics[width=0.4\textwidth]{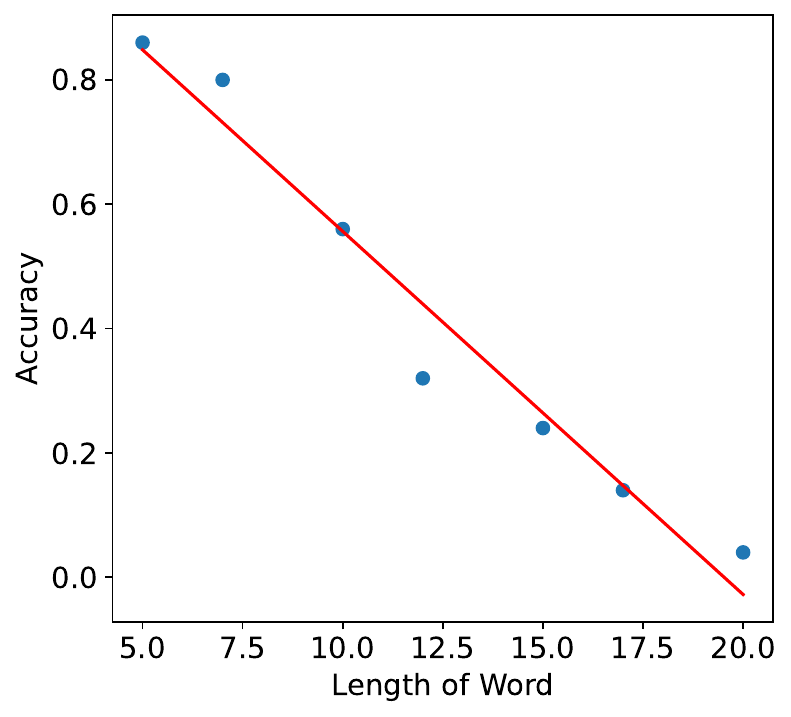}
    }
    \caption{Performance trends for the Two Sum and Word Search tasks. Both tasks exhibit a linear decline in accuracy as reasoning complexity increases: Two Sum with respect to the logarithm of array length, and Word Search with respect to word length.}
    \label{fig:case_study_trends}
\end{figure*}

To better understand how reasoning models fail on core operations, we analyze QwQ by combining controlled case studies with a general error analysis across the benchmark. The case studies target retrieval and backtracking with tasks whose operations are as simple as possible, so that the influence of other capabilities can be isolated. The general analysis further examines whether these failure patterns appear broadly across diverse tasks.

\paragraph{Retrieval Failures.}
We use Two Sum to test retrieval during reasoning. The problem asks the model to return the indices of two numbers in an array $nums$ that add up to a target. The reference solution iterates through the array and checks whether $target - nums[i]$ has appeared before. Solving this problem requires only two core capabilities: basic arithmetic (addition and subtraction) and retrieval.

We control the difficulty of the task by varying the length of the input array. \cref{fig:twosum} shows that accuracy decreases as the array length increases, exhibiting a linear relationship with the logarithm of the array length ($R^2 = 0.975$). Given that the expected number of reasoning steps scales linearly with array length, this trend aligns with the pattern observed in \cref{subfig:steps}. Notably, the model completely fails the task once the array length reaches 1,000.

We further inspect the model's reasoning traces and find that the errors are not caused by arithmetic mistakes, but by failures in basic retrieval operations. The retrieval errors mainly fall into two categories: the model may identify the correct value pair but retrieve incorrect indices, or it may fail to recognize that a correctly computed complement has already appeared in the array. We provide the detailed analysis in \cref{sec:retrieval_error_types}.

\paragraph{Backtracking Failures.}
We use Word Search to test backtracking during reasoning. Given an $m \times n$ grid of characters and a word, the model must return the list of positions that form the word in order. The word can be constructed from horizontally or vertically adjacent cells, and the same cell may not be used more than once. Solving this task requires only depth-first search (DFS) and does not involve any arithmetic computation.

We keep the size of the board fixed and vary the length of the word to control the depth of the DFS. As shown in \cref{fig:word}, the accuracy declines approximately linearly with increasing word length ($R^2=0.961$). When the required search depth reaches 20, the model struggles to complete the task successfully. By analyzing models' reasoning processes, we find that reasoning models exhibit limited backtracking capabilities. They are ineffective at exploring new paths and tend to repeatedly revisit previously explored ones. As illustrated in the example in \cref{sec:example}, the model repeatedly checks the sequences (3,10) → (3,9) → (2,9) → (2,8) → (1,8) → (0,8) → (0,7) → (0,6) and (3,10) → (3,9) → (4,9) → (4,8), but fails to explore further. We provide the detailed analysis in \cref{sec:example}.

\paragraph{General Error Analysis.}
To examine whether these patterns extend beyond the two case studies, we use GPT-5 \cite{gpt5_release} to inspect QwQ's reasoning traces where QwQ produces an incorrect final answer. We identify a retrieval failure when the model forgets information or conclusions from earlier in the reasoning process, or incorrectly reads the initial inputs or preceding context. We identify a backtracking failure when the model fails to adequately explore alternative possibilities, or becomes stuck repeatedly checking a specific reasoning segment without meaningful modification. The analysis shows that 71.1\% of incorrect samples exhibit retrieval failures, and 89.2\% exhibit backtracking failures. These results suggest that retrieval and backtracking failures are common challenges across a wide variety of tasks.

Overall, simple operations such as checking whether a number has appeared or determining whether a path has already been explored are typically not difficult for current models when they appear as isolated operations over a long input. However, models often fail when these operations must be performed repeatedly throughout a long reasoning process while maintaining and updating intermediate state. This distinction highlights why long reasoning introduces challenges that are not captured by long-input comprehension alone.

\section{Conclusion}
We present \ours{}, a benchmark designed to evaluate the long reasoning capabilities of large language models through algorithmic execution. \ours{} bridges a critical gap left by existing benchmarks focused primarily on long inputs. Our analysis reveals that model performance degrades predictably with increased required steps, and that current models struggle with basic retrieval and backtracking operations. \ours{} thus provides a scalable and controllable framework for probing and advancing the frontiers of large reasoning models. For future research, we will further broaden the evaluation of models' reasoning capabilities to include more diverse scenarios such as multimodal reasoning.

\section*{Limitations}
\label{sec:limitations}
To estimate the number of reasoning steps required for each task, we approximate it using the number of execution lines measured when running the solution code on the given input. While this generally reflects the reasoning complexity of the task, it may deviate from the exact number of steps required by the model, as the model may adopt different strategies to solve the problem. Nevertheless, precisely calculating the number of required reasoning steps for all tasks is infeasible, and our method serves as a sufficiently effective approximation. 

Furthermore, evaluation through algorithmic execution primarily focuses on deductive reasoning. It cannot cover all forms of reasoning, such as inductive reasoning or analogical reasoning.  Evaluating other types of reasoning remains an important direction for future work.

\section*{Broader Impacts}
\label{sec:impacts}
\ours{} focuses on evaluating models’ long reasoning capabilities, which facilitates the extension of their reasoning processes and the enhancement of their reasoning abilities. This leads to greater potential for solving complex real-world problems. However, as reasoning processes become longer and more sophisticated, it also becomes increasingly difficult to interpret and monitor the model’s reasoning trajectory, especially in cases involving harmful or deceptive intent.

\section*{Acknowledgments}

This work was supported by the Fundamental and Interdisciplinary Disciplines Breakthrough Plan of the Ministry of Education of China under Grant JYB2025XDXM504.

\bibliography{custom}

\appendix

\section{Inference Settings}

In this section, we detail the inference configurations used in our experiments. We employ two primary methods for model inference: local deployment via vLLM and remote access via APIs. For open-weights models served by vLLM, we apply specific generation parameters tailored to reasoning and non-reasoning models, as listed in \cref{tab:vllm_settings}. For proprietary models, we use the default API settings unless otherwise specified.

\label{sec:param}

\begin{table*}[t]
\centering
\begin{tabular}{@{}lcc@{}}
\toprule
              & \multicolumn{1}{l}{Reasoning Models} & \multicolumn{1}{l}{Non-Reasoning Models} \\ \midrule
MaxNewTokens  & 32K                                  & 8K                                       \\
System Prompt & N/A                                    & You are a helpful assistant            \\
Temperature   & \multicolumn{2}{c}{0.6}                                                         \\
TopP          & \multicolumn{2}{c}{0.95}                                                        \\
MinP          & \multicolumn{2}{c}{0}                                                           \\
TopK          & \multicolumn{2}{c}{40}                                                          \\
vLLM Version  & \multicolumn{2}{c}{0.8.5} \\
\bottomrule
\end{tabular}
\caption{Settings for vLLM Inference}
\label{tab:vllm_settings}
\end{table*}

\section{Evaluation Variance under Different Random Seeds}
\label{sec:var}

To assess the impact of random seeds on evaluation results, we re-evaluated Qwen3-8B on Level 1 and Level 2 using eight different seeds for each level. As shown in \cref{tab:var}, the mean accuracies are 49.1 and 15.4, with standard deviations of 0.86 and 0.65 for Level 1 and Level 2, respectively. The corresponding scores range from 47.8 to 50.6 and from 14.4 to 16.3, indicating that the evaluation results are stable across different random seeds.

\begin{table}[H]
\centering
\begin{tabular}{@{}lcccc@{}}
\toprule
Seed     & 1    & 2    & 3    & 4    \\ \midrule
Level 1  & 48.6 & 49.6 & 49.3 & 48.3 \\
Level 2  & 14.4 & 15.6 & 16.0 & 15.2 \\ \midrule
Seed     & 5    & 6    & 7    & 8    \\ \midrule
Level 1  & 50.6 & 49.3 & 49.0 & 47.8 \\
Level 2  & 14.6 & 15.3 & 16.3 & 15.7 \\
\bottomrule
\end{tabular}
\caption{Evaluation results of Qwen3-8B on Level 1 and Level 2 under different random seeds.}
\label{tab:var}
\end{table}

\section{Human Baseline}
\label{sec:human}

We conducted a preliminary human baseline evaluation on a randomly selected subset of our benchmark at Level 1 difficulty. The participants were seven computer science students, who participated on a strictly voluntary basis. Participants acknowledged their understanding of the data usage and agreed to participate by proceeding with the questionnaire. Prior to the task, participants were presented with specific guidelines to ensure standardization. The full text of the instructions provided was: "You are permitted to search for solutions and use computational tools; however, please do not directly execute code to solve the problem as a whole. Please record the time taken to complete the task." Following these instructions, the specific problem statements and their corresponding inputs were displayed to the participants.

In total, 20 questions were evaluated. On average, each question took 4 minutes to complete, and the participants achieved an average accuracy of 75\%. This result indicates that our benchmark poses a challenge not only for LLMs but also for human participants with relevant domain knowledge.

\section{Benchmark Construction Details}
\label{sec:construction_details}

\subsection{Tag Filtering}

We exclude problems tagged with categories such as database, pandas, linked list, multithreading, probability and statistics, randomized, design and interactive. Certain tags like database and linked list primarily test the use of libraries or implementation details, which is not the focus to our evaluation. Problems involving randomized behavior and multithreading are excluded because current LLMs are not well-suited for simulating these behaviors. In addition, since our benchmark is designed for single-turn interactions, we exclude interactive problems that require multi-turn or dynamic input/output exchanges.

\subsection{Core Problems}

\label{sec:core_problems}

To validate whether models exhibit more desired cognitive behaviors in filtered core problems, we segmented model’s reasoning traces and used GPT-4o to assess the presence of backtracking and memory read/write behaviors in each segment. Our analysis revealed statistically significant patterns:

\begin{itemize}
    \item Problems tagged with DFS/backtracking showed a higher frequency of backtracking behaviors in the traces (40.9\% vs. 32.4\%, p < 1e-4).
    \item Problems tagged with BFS/DP showed a higher frequency of memory-related behaviors (87.2\% vs. 77.2\%, p < 1e-4). 
\end{itemize}

These results suggest that the chosen tags are not arbitrary, but indeed correspond to different reasoning dynamics exhibited by the models.

\subsection{Code Coverage}

We used Python's coverage library to measure line-level code coverage. Specifically, we tracked the code coverage of the solution code during execution with ten different inputs (excluding import statements as well as class and function definitions). Below is an example of insufficient coverage: due to a bad input generator, all generated samples for this problem exit directly at the first return statement.

\begin{lstlisting}[language=Python]
class Solution:
   def __main(self, grid: List[List[int]]) -> int:
        if grid[0][1] > 1 and grid[1][0] > 1:
            return -1

        m, n = len(grid), len(grid[0])
        dist = [[inf] * n for _ in range(m)]
        dist[0][0] = 0
        q = [(0, 0, 0)]
        dirs = (-1, 0, 1, 0, -1)
        
        ......
\end{lstlisting}

\subsection{Execution Line Measurement}

The number of execution lines is the total number of source-code lines in the ground-truth solution that are actively executed when running a specific input. This count is dynamic: if a three-line code block inside a loop is executed ten times, it contributes 30 execution lines. We measure this metric programmatically rather than using an LLM or human annotation. Specifically, we use Python's \texttt{sys.settrace} function to track line-execution events at runtime while executing the solution code on each generated input. This produces one execution-line count for each problem-input pair.

\subsection{Sample Filtering}

For each problem with a qualified input generator, we generate 100 distinct input instances. Each generated input is paired with the corresponding problem statement and reference answer, where the reference answer is obtained by executing the ground-truth solution code.

We then filter the generated problem-input pairs before constructing the final test set. First, we remove inputs that are invalid, trigger execution errors, or exceed the input-length constraint. Second, we compute the execution-line count for each remaining input and retain samples that fall into the target difficulty ranges defined in the main text. Third, we apply model-based filtering to remove samples or problems that can be solved too easily through shortcuts or guessing. The final benchmark is sampled from the filtered pool. In \cref{tab:stat}, ``\# Problems'' denotes the number of unique algorithmic problems, whereas ``\# Samples'' denotes the number of distinct problem-input pairs. Since a single problem can be paired with multiple generated inputs, the number of samples naturally exceeds the number of problems.

\subsection{Case Study Input Generation}

For the Two Sum and Word Search case studies, we follow the same general input-generation principle but create problem-specific generators with controllable parameters. These generators are produced with LLM assistance and allow us to vary the scale of the inputs, such as the array length in Two Sum or the word length in Word Search. This controllability enables a more fine-grained analysis of retrieval and backtracking failures. The prompts given to the evaluated LLMs and the evaluation methodology are otherwise identical to those used in the main benchmark.

\subsection{Prompt}

\begin{tcolorbox}[breakable, title=Prompt for Evaluation]
Solve the given problem based on the given inputs.

Don't just reply with code. You should calculate the final answer step by step. Put your final answer within \textbackslash boxed\{\}.

\vspace{1\baselineskip}

Problem:

\{problem\}

\vspace{1\baselineskip}

Inputs:

\{inputs\}
\end{tcolorbox}

\begin{tcolorbox}[breakable, title=Prompt for Input Generator]
Write a Python function called "generate()" to generate test data for a given function. Follow these requirements:

1. Each run of the generate() function should return only one random data sample.

2. The return value of generate() must match the expected input format of the target function.

3. The format and range of the generated test data must follow the specifications in the Constraints section.

4. For any length-related quantities (e.g. list sizes, string lengths, number of elements), values should be sampled uniformly on a logarithmic scale within the allowed range. 

Notice to avoid undefined behavior with log(0), the lower bound must be at least log(1).

\vspace{1\baselineskip}

The code block should only include the "generate" function. Think step by step in the comments before the code.

\vspace{1\baselineskip}

Problem:

\{problem\}

\vspace{1\baselineskip}

Constraints:

\{constraints\}

\vspace{1\baselineskip}

Examples:

\{examples\}

\vspace{1\baselineskip}

Function:

\{solution code\}
\end{tcolorbox}

\begin{tcolorbox}[breakable, title=Prompt for Output Verifier]
You are a professional Python developer. Complete the following tasks:

**Task 1: Determine if the answer should be "order-independent"** 

I will provide a programming problem statement. Your task is to analyze whether the expected answer should be considered **order-independent**. For example the problem requires returning a **set**, a **list where answer could be in any order**, or a **combination or permutations of elements**.  

After analyzing the problem, output either `<flag>Order-dependent</flag>"` or `"<flag>Order-independent</flag>"`

\vspace{1\baselineskip}

**Task 2: Generate a Boolean Comparison Function**  

Based on the result of Task 1, generate a Python function 'compare(user\_output, expected\_output)', which compares the output of a user's submission ('user\_output') with the expected answer ('expected\_output') and **returns a boolean value ('True' or 'False')**: 

- **If the answer is order-independent**, the function should compare the outputs while ignoring order.  

- **If the answer is order-dependent**, the function should compare the outputs strictly as given.  

The code block should only include the "compare" function.

\vspace{1\baselineskip}

Problem:

\{problem\}

\vspace{1\baselineskip}

Examples:

\{examples\}
\end{tcolorbox}

\begin{figure*}[h]
    \centering
    \subfigure[]{
        \includegraphics[width=0.45\textwidth]{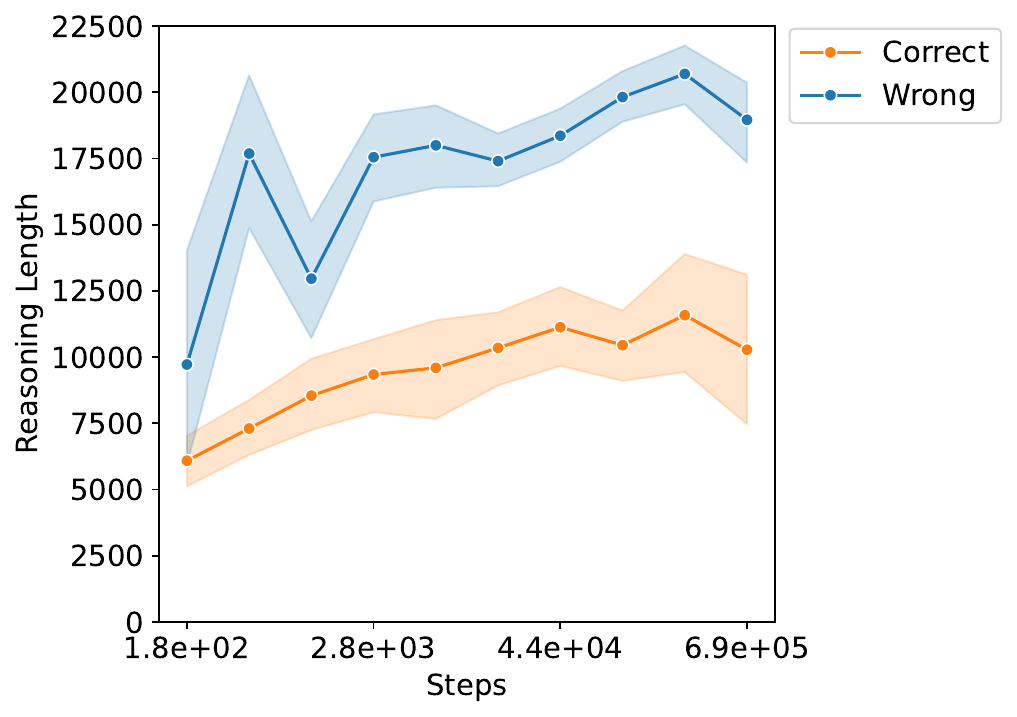}
    }
    \subfigure[]{
        \includegraphics[width=0.45\textwidth]{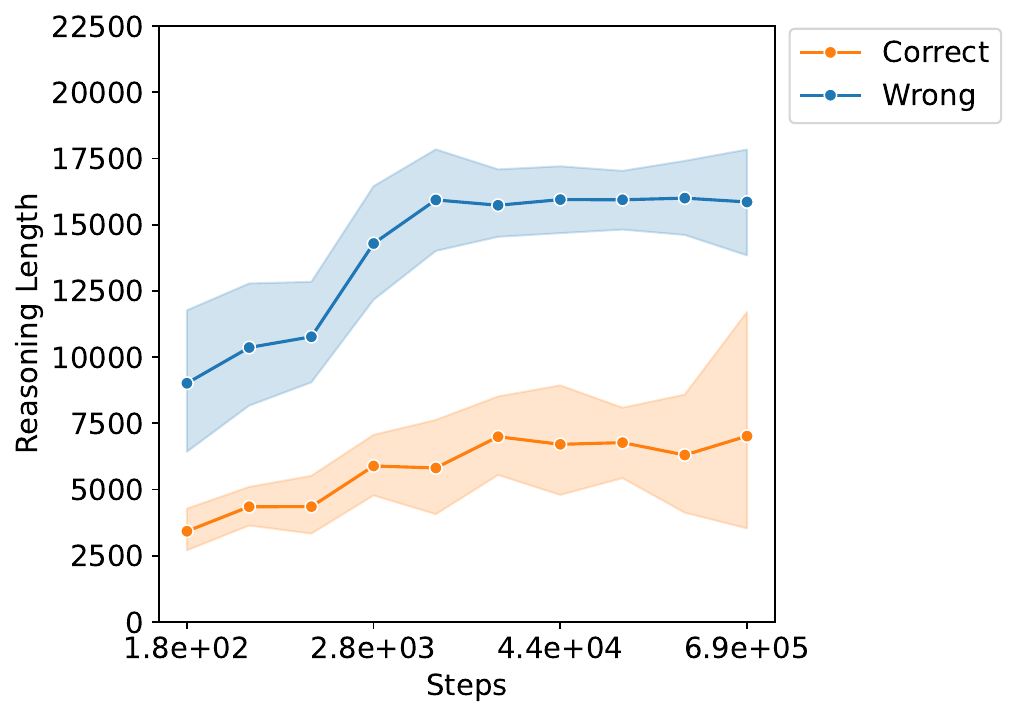}
    }
    
    \caption{Relationship between the expected number of reasoning steps and reasoning length of
    (a) QwQ and (b) DeepSeek-R1-Distill-Qwen-32B. The reasoning length of correct samples generally increases with higher reasoning complexity, but the reasoning length of incorrect samples is significantly longer than that of correct ones}
    \label{fig:thought}
\end{figure*}

\section{Retrieval Failure Analysis}
\label{sec:retrieval_error_types}

To determine whether the primary source of error lies in arithmetic computation or retrieval, we extract equations from the model's reasoning process and evaluate their correctness. The correctness rate of these equations reaches 98\%. Moreover, for this task, the computations at different indices are independent, and the model only needs to perform the correct computation at the index corresponding to the answer in order to succeed. Therefore, there is no accumulation of errors across steps. These findings indicate that the dominant source of failure lies in the model's retrieval ability.

Further examination of erroneous Two Sum cases reveals that errors can be categorized into index errors and full errors. An index error occurs when the model identifies the correct value pair but retrieves incorrect indices. In the following example, given the input $nums[74] = -5377$, $nums[75] = -22651$, $nums[76] = 27401$, the model correctly computes based on $nums[74]$, but fails to retrieve $nums[75]$. Instead, it mistakenly assigns the value of $nums[76]$ to index 75. If the skipped value is not part of the correct answer, the model may still find the correct value pair; however, this leads to systematic misalignment in all subsequent indexes.

\begin{tcolorbox}[breakable]
Index74: -5377 → complement - 22316 → add.

Index75: 27401 → complement - 55094 → add.
\end{tcolorbox}

A full error refers to a failure to identify the correct value pair. In the following example, the correct answer is [156, 382] ($target = 1973$ , $nums[156] = 31753$ , $nums[382] = -29780$). Although the model correctly computes that the number corresponding to -29,780 is 31,753, it fails to retrieve that 31,753 has already appeared in the input.

\begin{tcolorbox}[breakable]
Index156:31753 → complement is negative. Not present.

...

Index382:-29780 → complement 1973 + 29780 = 31753. Not present.
\end{tcolorbox}

\begin{figure}[h]
    \centering
    \includegraphics[width=0.45\textwidth]{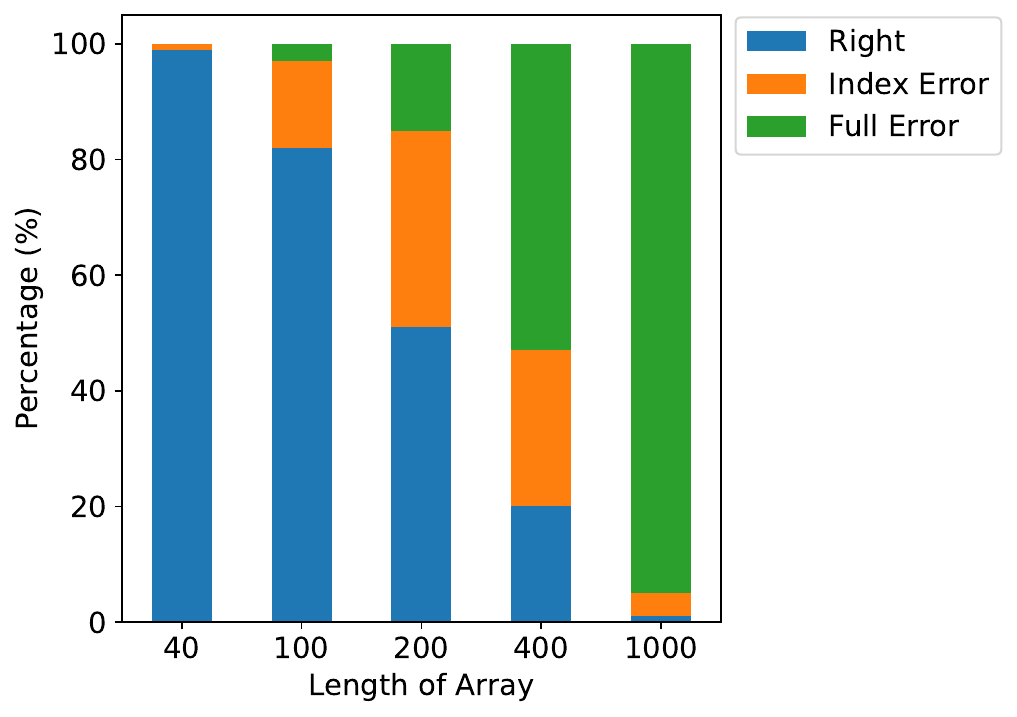}
    \caption{Error type distribution for the Two Sum task. The distribution shows a transition from index errors to full errors as array length increases.}
    \label{fig:twosum_errors}
\end{figure}

As shown in \cref{fig:twosum_errors}, when the array length is relatively short, most errors are index errors. However, as the array length increases, the proportion of full errors rises steadily and eventually becomes the dominant error type. This further demonstrates that as the array length grows, the retrieval task becomes increasingly difficult for the model.

\section{Backtracking Failure Example}
\label{sec:example}

To quantify the breadth of exploration, we define a valid path as a cell sequence $S = {(x_i, y_i)}$ of length greater than two, where each adjacent pair of cells is spatially adjacent and satisfies $board[x_i, y_i] = word[i]$. Furthermore, if a valid path is not a subpath of any other valid path, we classify it as a distinct path. We measure the number of distinct paths discovered during reasoning process and observe that the model discovers only 4.9 distinct paths on average. This small number indicates insufficient exploration. In the example below, the model repeatedly checks previously explored paths but fails to backtrack to the branch that leads to the correct answer.

\begin{tcolorbox}[breakable]
So moving left to (4,8). Now, the path is (3,10) → (3,9) → (4,9) → (4,8). 

...

So path now is (3,10) → (3,9) → (2,9) → (2,8) → (1,8) → (1,7). 

...

Path so far: (3,10) → (3,9) → (2,9) → (1,9). 

...

Now the path is (3,10) → (3,9) → (4,9) → (4,8) (J). 

...

1. (3,10) \myline 2. (3,9) \myline 3. (2,9) \myline Now, third letter is the second K.

<omit intermediate steps>

Next is U. \myline Left to (0,6) (U) \myline

...

Wait, let me try starting at (3,10) again but take a different route after the two Ks. \myline Path so far: \myline 1. (3,10) T \myline 2. (3,9) K \myline 3. (4,9) K \myline Now, next is J. \myline From (4,9), move left to (4,8) J. 

...

(3,10) → (3,9) → (2,9) → (2,8) → (1,8) → (0,8) → (0,7) → (0,6) → then stuck.

...

Wait let me try one more time with the path starting at (3,10) and see if there's another route. \myline Path: \myline 1. (3,10) T \myline 2. (3,9) K \myline 3. (4,9) K 

...

Suppose the path starts at (3,10) → (3,9) K → then to (2,9) K → then to (2,8) J → then to (1,8) H → then to (0,8) Q → (0,7) T → (0,6) U → then need M. 

...

1. (3,10) T \myline 2. (3,9) K \myline 3. (4,9) K \myline 4. (4,8) J 
\end{tcolorbox}

\section{Relationship Between Reasoning Length and Complexity}

We analyze the relationship between the model’s reasoning length and the reasoning complexity of tasks. As shown in \cref{fig:thought}, the reasoning length of correct samples generally increases as the number of required steps increases. This indicates that the model is, to some extent, able to dynamically adjust its reasoning length according to task complexity. However, it is noteworthy that the reasoning length of incorrect samples is significantly longer than that of correct ones. This suggests that the model could not fully leverage its extended reasoning process—longer reasoning does not guarantee better problem-solving. This observation is also consistent with findings from prior work\cite{underthinking}. 

To further understand why failed cases tend to produce longer reasoning chains, we compare the reasoning strategies in successful and failed cases. Qualitatively, failed traces often exhibit oscillation between incorrect results: the model may move from one erroneous answer to another, revisit a previously rejected result, and repeat similar checks without making substantive progress. This behavior differs from useful self-correction, where the model identifies a concrete mistake and then proceeds toward a corrected solution. Such harmful oscillation consumes many tokens and can derail the model into logical confusion, whereas successful cases usually follow a more direct path to the correct solution.

To quantify this behavioral difference, we use GPT-5 to analyze QwQ's reasoning traces on Level 1 problems. Specifically, we prompt GPT-5 to count oscillations in each trace and distinguish harmful oscillation from genuine self-correction. The results show that successful cases contain only 0.2 harmful oscillations per trace on average, whereas failed cases contain 1.03. This contrast helps explain why failed cases have substantially longer reasoning chains: a considerable portion of the additional reasoning length is spent revisiting unstable or incorrect hypotheses rather than making productive progress.

\section{Summary of Linear Regression Analyses}
\label{sec:linear}

\begin{table}[h]
\centering
\begin{tabular}{@{}lcc@{}}
\toprule
                  & $R^2$ & p-value         \\ \midrule
GPT-5                & 0.964  & \textless{}1e-4 \\ 
QwQ               & 0.930  & \textless{}1e-4 \\ 
Claude Opus 4.5 & 0.964  & \textless{}1e-4 \\ 
o3       & 0.966  & \textless{}1e-4 \\ \bottomrule
\end{tabular}
\caption{The significance of linear regression between model accuracy and the logarithm of expected number of reasoning steps.}
\end{table}

\section{The Use of Large Language Models}
In paper writing, we use Large Language Models to polish writing. In experiments, we use Large Language Models both as our research subjects and as tools to generate training data.

\section{License of Artifacts}

The LeetCode data is licensed under CC-BY-SA-4.0. Furthermore, all models utilized and evaluated are confirmed to be used in accordance with their respective terms of use. Our benchmark is intended to be used for academic research purposes.

\end{document}